\documentclass[letterpaper, 10 pt, conference]{ieeeconf}  

\IEEEoverridecommandlockouts                              

\usepackage{amsmath,amssymb,amsfonts}
\usepackage{graphicx}
\usepackage{textcomp}
\usepackage[english,ruled,linesnumbered]{algorithm2e}
\usepackage{lipsum}
\usepackage{array}
\usepackage{subfigure}
\usepackage{multirow}
\usepackage{color} 
\usepackage{threeparttable}
\usepackage{cite}

\title{\LARGE \bf
Topological Semantic Mapping by Consolidation of Deep Visual Features
}

\author{Ygor C. N. Sousa and Hansenclever F. Bassani
\thanks{This work was supported by the Brazilian Coordination for Improvement of Higher Education Personnel (CAPES).}
\thanks{The authors are with the Centro de Informatica,
        Universidade Federal de Pernambuco, Recife, PE, 50.740-560, Brazil
        {\tt\small \{ycns, hfb\}@cin.ufpe.br}}%
}

\begin{document}

\maketitle
\thispagestyle{empty}
\pagestyle{empty}

\begin{abstract}

Many works in the recent literature introduce semantic mapping methods that use CNNs (Convolutional Neural Networks) to recognize semantic properties in images. The types of properties (eg.: room size, place category, and objects) and their classes (eg.: kitchen and bathroom, for place category) are usually predefined and restricted to a specific task. Thus, all the visual data acquired and processed during the construction of the maps are lost and only the recognized semantic properties remain on the maps. 
In contrast, this work introduces a topological semantic mapping method that uses deep visual features extracted by a CNN (GoogLeNet), from 2D images captured in multiple views of the environment as the robot operates, to create, through averages, consolidated representations of the visual features acquired in the regions covered by each topological node. These representations allow flexible recognition of semantic properties of the regions and use in other visual tasks. Experiments with a real-world indoor dataset showed that the method is able to consolidate the visual features of regions and use them to recognize objects and place categories as semantic properties, and to indicate the topological location of images, with very promising results. 

\end{abstract}

\section{INTRODUCTION}

The semantic mapping process involves using machine learning methods to infer semantic properties (eg.: room size, place category, and objects) of the mapped environments at a metric or topological level \cite{kostavelis2015,pronobis2012}. Many methods proposed in recent works use CNNs (Convolutional Neural Networks) to recognize semantic properties through classification, object detection and semantic segmentation in images and integrate the results with metric or topological maps in several different approaches \cite{ma2017,xiang2017,sunderhauf2016,roddick2020,mccormac2017,maturana2017,maturana2018,bernuy2018,luo2018, sunderhauf2017,rangel2019,sousa2018,nakajima2019,rosinol2021}.

Despite the benefits of using CNNs, training this type of neural network is costly in terms of time and resources. Using pre-trained CNNs is an alternative already present in the recent semantic mapping literature. Some works use a pre-trained CNN and perform transfer learning, fine-tuning it to recognize task-specific semantic information in the images \cite{sunderhauf2016,mccormac2017,xiang2017,roddick2020}. Others use a pre-trained CNN to perform the recognition of semantic properties without any changes or retraining \cite{sunderhauf2017,rangel2019, sousa2018,nakajima2019,luo2018,rosinol2021}.

Most semantic mapping methods attach to the maps single-view semantic information recognized from images with semantic data accumulated or fused over time \cite{mccormac2017,maturana2017,sunderhauf2017, maturana2018,bernuy2018,luo2018,rangel2019}. Other methods change the common approach of working only in single-view and acquire semantic information by performing multi-view semantic segmentation \cite{ma2017,xiang2017}. However, the types of semantic properties recognized are usually previously defined and thus restricted to what was planed. Due to these restrictions, the large amount of the visual data acquired and processed by the agent during the building process of the maps is lost, no new semantic recognition can be made from them, and only the recognized semantic properties remain on the maps.

The present work introduces a topological semantic mapping method that uses deep visual features extracted from 2D images acquired in multiple perspectives of the environment, as the robotic agent operates, to create representations averaging the visual features extracted from the images observed by the robot around each topological node. These representations, denoted as ``consolidated representations'', allow flexible recognition of semantic properties of the regions and use in other types of visual tasks, such as visual localization on the map and visual search of object instances. The features are extracted from a deep latent layer of a pre-trained CNN, the GoogLeNet (Inception v1) \cite{szegedy2015}, and consolidated into the nodes of a topological map in a process inspired by Self-Organizing Maps (SOM) \cite{bassani2015}. In this work, the consolidated representations are used to perform the classification of objects and place categories as semantic properties of the regions covered by the nodes. The objects are classified using the classification layer of GoogLeNet, without retraining, and the place categories are recognized using a shallow MLP (Multilayer Perceptron).

The method was evaluated in experiments using the COLD dataset (COsy Localization Database)~\cite{pronobis2009}, a real-world indoor dataset. We show it can build proper maps and consolidate visual features used to recognize semantic properties (objects and place categories) of regions, and to indicate the topological location of images, with very promising results. The experiments also suggested that the consolidated representations of visual features did not degrade over time. Furthermore, the approach used to locate images on the maps is also introduced in this work.

The main contributions of this work are:

\begin{itemize}
    \item{A topological mapping method with consolidation of deep visual features through averages, visual persistence between nodes, and visual habituation;}
    \item{We demonstrate the richness of the consolidated features by showing that they can be used for extracting different semantic properties even defined after the map is built;}
    \item{An approach based on adaptive Euclidean distances to indicate the location of images on topological maps using the consolidated features.}
\end{itemize}

\section{RELATED WORK}
\label{sec:relatedwork}

Many works in the literature of semantic mapping use CNNs to recognize semantic properties in various types of approaches and applications. In \cite{mccormac2017}, McCormac et al. use a CNN to perform semantic segmentation of images and incorporate the predictions into a dense 3D metric map. In a method created for Micro-Aerial Vehicles (MAV), Maturana et al. \cite{maturana2017} present a CNN designed for fast on-board processing. The CNN performs semantic segmentation of images and the 2D measurements are aggregated into a 2.5D grid map. In other work, Maturana et al. \cite{maturana2018} propose a CNN designed to perform semantic segmentation and incorporate the semantic information into a 2.5D grid map for off-road autonomous driving. Bernuy and Ruiz-del-Solar \cite{bernuy2018} use a CNN to perform semantic segmentation of images and the output labels are accumulated in histograms that are used to assist the creation of topological maps.

In a hybrid metric-topological semantic mapping approach, Lou and Chiou \cite{luo2018} use a CNN to detect objects in a mapping system with multiple levels of semantic information. In \cite{sunderhauf2017}, Sunderhauf et al. detect objects in images using a pre-trained CNN and integrate the results into a 3D sparse metric map, where a 3D unsupervised segmentation method is used to assign segments of 3D points of the map to the detected objects. Nakajima and Saito \cite{nakajima2019} use a pre-trained CNN to detect objects in images and improve 3D maps with object-oriented semantic information in an efficient framework.

The method introduced by Xiang and Fox \cite{xiang2017} presents a recurrent approach that accumulates features extracted with a CNN to perform multi-view semantic segmentation in a joint 3D scene semantic mapping process. Despite advances in the method that allow the accumulation of features from multiple views, it does not create representations of regions that can be used to obtain other semantic information and the accumulated features are only used to label pixels of images.

Some works use pre-trained CNNs integrated with other machine learning approaches to allow incremental learning of new classes of a semantic property. In \cite{sunderhauf2016}, Sunderhauf et al. introduce a model that learns incrementally new categories of places, in a supervised fashion, by extending a CNN with an one-vs-all Random Forest classifier. Rangel et al. \cite{rangel2019} present a semantic mapping method that exploits CNNs previously trained to classify unlabeled images and perform a bottom-up aggregation approach that clusters images in the same semantic category. In \cite{sousa2018}, Sousa and Bassani presented a topological semantic mapping approach that incrementally and in an on-line fashion, forms clusters of object vectors classified with a pre-trained CNN. The clusters are formed using a SOM-based method and represent the categories of visited places. However, even with the efforts to allow semantic mapping methods to incorporate new semantic classes incrementally, these methods are still restricted to previously defined types of semantic properties. 

In contrast, this work builds consolidated representations of visual features from the regions delimited by the topological nodes, and the desired semantic information can be obtained from these representations using a wide variety of machine learning methods. This presents an advantage over previous methods as the consolidated representations are rich and allow new semantic information to be obtained whenever necessary, increasing the flexibility of the produced maps.

\section{THE METHOD}
\label{sec:method}

\begin{figure*}[ht]
\centering
  \includegraphics[width=0.98\textwidth]{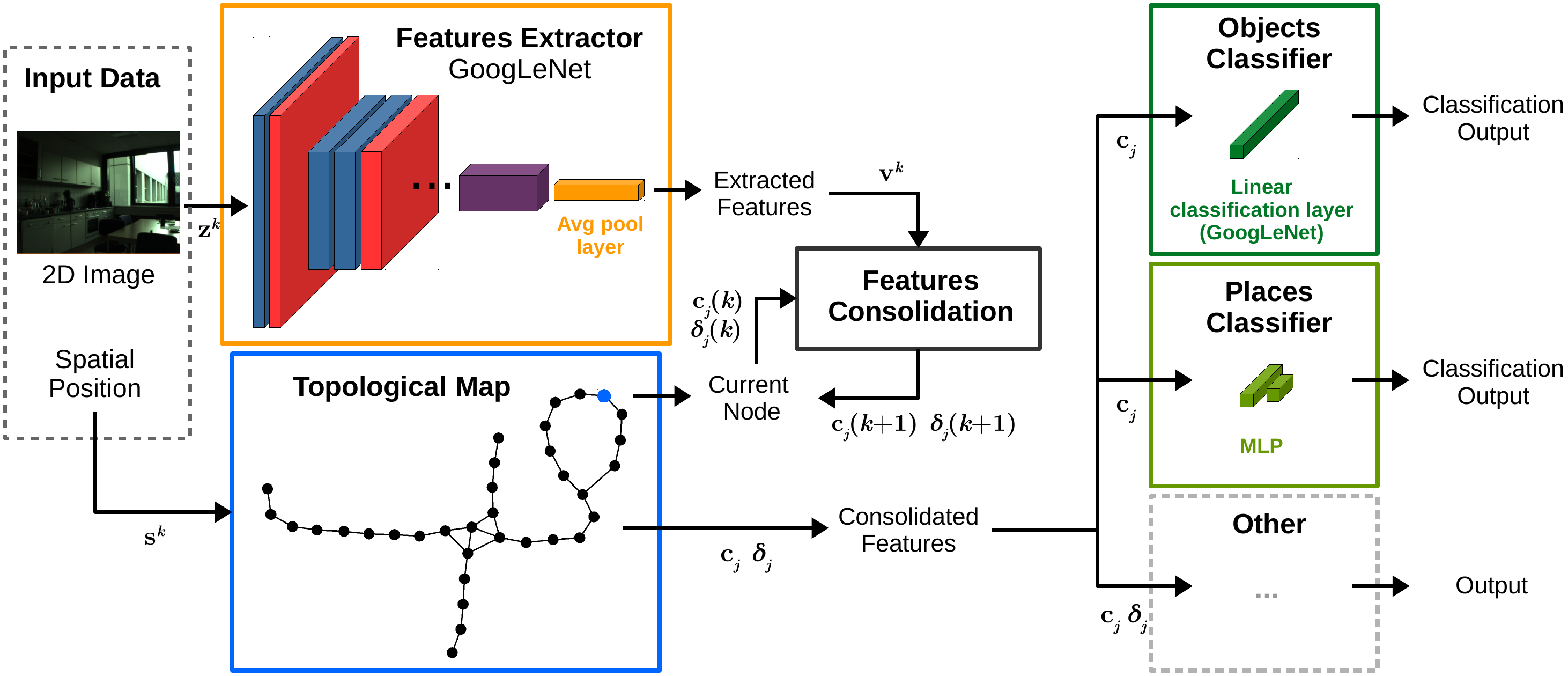}
  \caption{Overview of the method: GoogLeNet extracts the visual features vector $\mathbf{v}^{k}$ of the input images and the nodes on the topological map corresponding to the input positions consolidate $\mathbf{v}^{k}$ in their representations; The consolidated visual features vector $\mathbf{c}_{j}$ in each node is provided to the classification layer of the GoogLeNet (without retraining) and the objects visualized in the nodes are obtained; The vector $\mathbf{c}_{j}$ in each node is also provided to a shallow MLP and the place category of each node is obtained; The consolidated vectors $\mathbf{c}_{j}$ and $\pmb{\delta}_{j}$, the average features distance vector, can be used to recognize other semantic properties or in other visual tasks, as in the topological image localization experiments (Section \ref{sec:loc-evaluation}) which uses $\mathbf{c}_{j}$ and $\pmb{\delta}_{j}$.}
  \label{fig:overviewMethod}
  \vspace{-0.5cm}
\end{figure*}

The topological semantic mapping method presented in this work is inspired by SOM with time-varying structure \cite{araujo2013} and uses the GoogLeNet, pre-trained on ImageNet \cite{imagenet2009} and available on the PyTorch library, to extract deep visual features from 2D images. An overview of the method is illustrated in Fig. \ref{fig:overviewMethod} and the following sections describe in detail the operation of the method.

\subsection{Topological Mapping}
\label{sec:topology}

At each time step $k$, the method receives as input a 2D image $\mathbf{z}^{k}$ and its spatial position of capture $\mathbf{s}^{k}=\{s^{k}_{i},i=1...n\}$ in the environment, where $n = 2$ in this work, but it could easily be extended to $3$ dimensions for 3D maps. Each image is provided as input to the GoogLeNet, the output of the adaptive average pool 2D layer (input of the linear classification layer) is flattened in the vector $\mathbf{v}^{k}$ of $m=1024$ dimensions and used as a visual features vector.

The topological map starts empty and is built as a graph. Each node $j$ in the graph contains three vectors: $\mathbf{p}_{j}=\{p_{ji},i=1...n\}$, the spatial position of the node; $\mathbf{c}_{j}=\{c_{ji},i=1...m\}$, the consolidated visual features vector; and $\pmb{\delta}_{j}=\{\delta_{ji},i=1...m\}$, the average features distance vector.

As the input data ($\mathbf{v}^{k}$ and $\mathbf{s}^{k}$) is provided, the nodes compete to determine the topological location of the agent on the map and establish the nodes that will consolidate the visual features $\mathbf{v}^{k}$. The competition \cite{araujo2013} is performed using the spatial location provided, $\mathbf{s}^{k}$, and the winner is the node with the smallest spatial distance $D_{e}(\mathbf{s}^{k},\mathbf{p}_{j})$ to $\mathbf{s}^{k}$:
\begin{equation} \label{eq:vencedorCompeticaooTopologica}
h(\mathbf{s}^{k}) =  \operatorname*{arg\,min}_j [D_{e}(\mathbf{s}^{k},\mathbf{p}_{j})],
\end{equation}
where $D_{e}(\mathbf{s}^{k},\mathbf{p}_{j})$ is calculated as the Euclidean distance between the given spatial position $\mathbf{s}^{k}$ and the spatial position of the node $\mathbf{p}_{j}$:
\begin{equation} \label{eq:calculoDistanciaNodo}
D_{e}(\mathbf{s}^{k},\mathbf{p}_{j}) = \sqrt{\displaystyle\sum_{i=1}^{n} (s^{k}_{i}-p_{ji})^{2}}.
\end{equation}

If the map is empty or the spatial distance of the winning node to $\mathbf{s}^{k}$ is greater than the spatial distance threshold $\lambda$, a new node $\eta$ is inserted into the map with $\mathbf{p}_{\eta}=\mathbf{s}^{k}$, $\pmb{\delta}_{\eta} = \mathbf{0}$, and the consolidated visual features vector $\mathbf{c}_{\eta}$ initialized through an average between the input visual features $\mathbf{v}^{k}$ and the current state of the consolidated visual features vector $\mathbf{c}_{l}$ of the last visited node $l$, if any:
\begin{equation} \label{eq:featuresVectorInitialization}
\mathbf{c}_{\eta} = 
  \begin{cases}
  \displaystyle
     \gamma \mathbf{c}_{l} + (1 - \gamma) \mathbf{v}^{k}, & \text{if there is a }  \mathbf{c}_{l}\\
    \mathbf{v}^{k},   & \text{otherwise,}\\
  \end{cases}
\end{equation}
where $\gamma_{}$ is the persistence rate, which determines how much of the consolidated visual features vector of the last visited node will persist on the new node.

If the spatial distance of the winning node is equal to or smaller than the spatial distance threshold $\lambda$, the node consolidates the input visual features $\mathbf{v}^{k}$ as described in the next section (Section \ref{sec:visual-consolidation}). Moreover, if there are other nodes with spatial distance higher, but equal to or smaller than $\lambda$, these nodes also consolidate the input visual features $\mathbf{v}^{k}$. This is because an image may be captured by the agent at an intersection position between nodes.

A connection between nodes is created whenever a transition occurs between them as the agent moves through the environment, and a transition is determined when the consecutive winning nodes are different. In addition, when a new node is inserted into the map, a connection is created between the new node and the previous winner, if any.

\subsection{Visual Features Consolidation}
\label{sec:visual-consolidation}

The visual features in the input data are consolidated in the node $j$ by updating the vector $\mathbf{c}_{j}$ through a moving average considering the learning rate $\alpha_{}\in{]0,1[}$ and the squared Euclidean distance, $D_{}(\mathbf{v}^{k},\mathbf{u}_{j})$, between the visual features input vector $\mathbf{v}^{k}$ and the last visual features vector consolidated in node $j$, $\mathbf{u}_{j}$:
\begin{equation} \label{eq:featuresVectorConsolidation}
\mathbf{c}_{j} (k + 1) = 
  \begin{cases}
  \displaystyle
     \mathbf{c}_{j}(k) + \alpha (\mathbf{v}^{k} - \mathbf{c}_{j}(k)), & \text{if }  D_{}(\mathbf{v}^{k},\mathbf{u}_{j}) \geq \tau\\
    \mathbf{c}_{j}(k),   & \text{otherwise,}\\
  \end{cases}
\end{equation}
where $\tau$ is the minimum distance between the input features vector and the last features vector consolidated in the node, which prevents the node from consolidating very similar visual features vectors in sequence and, therefore, controls the visual habituation of the node. $D_{}(\mathbf{v}^{k},\mathbf{u}_{j})$ is the squared Euclidean distance described as follows:
\begin{equation} \label{eq:distanciaEuclidianaConsolidacao}
D_{}(\mathbf{v}^{k},\mathbf{u}_{j}) =  \displaystyle\sum_{i=1}^{m} (v^{k}_{i}-u_{ji})^{2}.
\end{equation}

Also through a moving average (introduced in \cite{bassani2015}), the average features distance vector $\pmb{\delta}_{j}$ of node $j$ is updated considering the distance between $\mathbf{v}^{k}$ and $\mathbf{c}_{j}$:
\begin{equation} \label{eq:atualizacaoDistanciaMedia}
\pmb{\delta}_{j}(k+1) = 
  \begin{cases}
  \displaystyle
     \pmb{\delta}_{j}(k) + \alpha\beta(\pmb{\phi} - \pmb{\delta}_{j}(k)), & \text{if }  D_{}(\mathbf{v}^{k},\mathbf{u}_{j}) \geq \tau\\
    \pmb{\delta}_{j}(k),   & \text{otherwise,}\\
  \end{cases}
\end{equation}
where $\beta \in{]0,1[}$ controls the rate of change of the moving average, $\alpha$ is the learning rate, $\pmb{\phi} = | \mathbf{v}^{k}- \mathbf{c}_{j}(k)|$ denotes the absolute value of each component of the resulting difference vector, and $\tau$ controls the visual habituation of the node. After the update, if $\mathbf{c}_{j}$ and $\pmb{\delta}_{j}$ were updated in relation to $\mathbf{v}^{k}$, then $\mathbf{u}_{j}=\mathbf{v}^{k}$.

\subsection{Object and Place Classification}
\label{sec:semantic-information}

In this work, the visual features consolidated in the vector $\mathbf{c}_{j}$ are used to classify the objects visualized by the agent in the regions covered by each node $j$. For that, the vector $\mathbf{c}_{j}$ of each node is provided to the linear classification layer of GoogLeNet and the classification outputs are obtained without the need of retraining.

In addition, the visual features consolidated in the vector $\mathbf{c}_{j}$ are also used to obtain the place categories of the regions covered by each node $j$. A simple MLP implemented with the PyTorch library classifies the vector $\mathbf{c}_{j}$ of each node and the place category is obtained. The MLP contains one hidden layer with 20 units and ReLU (Rectified Linear Unit) activation functions. Furthermore, the loss function used for training in the MLP was cross entropy and the optimizer was Adam \cite{kingma2014}. In the experiments described in Section \ref{sec:experiments}, the MLP is trained with visual features vectors of 1024 dimensions extracted from 2D images using the GoogLeNet, where the output of the adaptive average pool 2D layer is the visual features vector of each image used for training.

The classification of the objects and place category of each node can be performed whenever necessary with the current state of the vector $\mathbf{c}_{j}$. The same could be done to recognize other semantic properties of nodes using, for example, new classifiers through transfer learning or clustering methods. 

\subsection{Image Localization}
\label{sec:image-localization}

To present the use of a generated map in other type of visual task, we use the consolidated visual features vector $\mathbf{c}_{j}$ and the average features distance vector $\pmb{\delta}_{j}$ of each node $j$ to localize 2D images on the map. The images have their features extracted using GoogLeNet, where the output of the adaptive average pool 2D layer is used as the visual features vector $\mathbf{x}$ of each image. The node with the highest activation $ac(D_{\omega}(\mathbf{x},\mathbf{c}_{j}), \pmb{\omega}_{j})$ represents the estimated location of the image on the map:

\begin{equation} \label{eq:winnerCompetitionLocalization}
loc(\mathbf{x}) =  \operatorname*{arg\,max}_j [ac(D_{\omega}(\mathbf{x},\mathbf{c}_{j}), \pmb{\omega}_{j})].
\end{equation}

Thus, $\pmb{\omega}_{j}$ is the relevance vector of the node $j$, in which each component is computed by an inverse logistic function based on the one proposed on \cite{bassani2015}:
\begin{equation} \label{eq:relevanceVectorCalculation}
\omega_{ji} = 
  \begin{cases}
  \displaystyle
    \frac{1}{1+exp\Big(\small\frac{\delta_{ji}-\delta_{jimean}}{s(\delta_{jimax}-\delta_{jimin})}\Big)},   & \text{if }  \delta_{jimin} \neq \delta_{jimax}\\
    1,   & \text{otherwise,}\\
  \end{cases}
\end{equation}
where $\delta_{jimin}$, $\delta_{jimax}$ and $\delta_{jimean}$ are respectively the minimum value, the maximum value and the average of the components of the distance vector $\pmb{\delta}_{j}$. $s>0$ controls the slope of the sigmoid function.

The activation $ac(D_{\omega}(\mathbf{x},\mathbf{c}_{j}), \pmb{\omega}_{j})$ for each node $j$ is calculated by a function of the weighted distance and the sum of the components of the relevance vector (introduced on \cite{bassani2012}):
\begin{equation} \label{eq:calculoAtivacaoLocalizacao}
ac(D_{\omega}(\mathbf{x},\mathbf{c}_{j}), \pmb{\omega}_{j}) = \dfrac{\left\lVert \pmb{\omega}_{j}\right\rVert_1}{D_{\omega}(\mathbf{x},\mathbf{c}_{j}) + \left\lVert \pmb{\omega}_{j}\right\rVert_1 + \varepsilon},
\end{equation}
where $\varepsilon$ is a small value to avoid division by zero, $\left\lVert.\right\rVert_1$ is the $L^1$-norm, and $D_{\omega}(\mathbf{x},\mathbf{c}_{j})$ is the weighted Euclidean distance between the consolidated visual features vector and the image visual features vector, with weights given by $\pmb{\omega}_{j}$:
\begin{equation} \label{eq:distanciaPonderada}
D_{\omega}(\mathbf{x},\mathbf{c}_{j}) =  \sqrt{\displaystyle\sum_{i=1}^{m} \omega_{ji}(x_{i}-c_{ji})^{2}}.
\end{equation}

Such an adaptive distance is validated in \cite{bassani2015} and provides better results with high dimensional data. Furthermore, in this work we use $\pmb{\delta}_{j}$ only for the task of topological image localization, but we will demonstrate in future work its use to obtain semantic properties through unsupervised and semi-supervised subspace clustering methods. 

\section{EXPERIMENTS}
\label{sec:experiments}

In this section, we detail the experiments performed to evaluate the topological maps produced with the proposed method and their consolidated representations of deep visual features. The quality of the consolidated representations is evaluated in the tasks of object classification, place classification, and topological image localization.

\subsection{Setup}
\label{sec:setup}

We used in the experiments the COLD dataset, a real-world indoor dataset that has three sub-datasets acquired in three different laboratories, COLD-Freiburg, COLD-Ljubljana, and COLD-Saarbrucken. Each sub-dataset contains data sequences acquired in different paths in the facilities. Not all data sequences in the dataset were used due to inaccuracies in the position data provided. We selected for use in experimentation 18 data sequences of 6 paths, 3 sequences of each path. Of the total, 6 data sequences of 2 paths are from COLD-Freiburg, where path 1 is a sub-part of path 2. In addition, 12 sequences of 4 paths are from COLD-Saarbrucken, in which paths 1 and 3 are, respectively, sub-parts of paths 2 and 4. The dataset has different types of data, but only 2D images from regular camera and their acquisition positions were used. The positions were computed by the dataset authors with a laser-based localization technique. In all selected data, there are 11 place categories, 12592 images in COLD-Freiburg and 17700 in COLD-Saarbrucken. 

Furthermore, we tuned almost all hyperparameters of the proposed method (and image localization approach) for the following experiments using a parameter sampling technique, the Latin Hypercube Sampling (LHS) \cite{Helton2005}. We set the spatial distance threshold $\lambda$ manually since it directly affects the topology of the maps and the value $0.9$ was used for all experiments. The LHS samples the hyperparameters within previously established ranges and the ranges of each hyperparameter are divided into subintervals of equal probability. Then, a single value is randomly chosen from each subinterval. Tab.~\ref{table:tableParamRanges} presents the ranges used for each hyperparameter.

\newcolumntype{C}[1]{>{\centering\arraybackslash}p{#1}}
\begin{table}[ht]
\centering
\caption{Hyperparameter ranges.}
\label{table:tableParamRanges}
\begin{tabular}{lcc}
\hline
&&\\[-2ex]
Hyperparameter & min & max \\[0.5ex]
\hline
&&\\[-2ex]
Features learning rate ($\alpha_{}$) & 0.001 & 0.1 \\[0.5ex]
Minimum distance ($\tau$) & 1.0 & 100.0\\[0.5ex]
Persistence rate ($\gamma$) & 0.001 & 0.999 \\[0.5ex]
Moving average rate ($\beta$) & 0.001 & 0.999 \\[0.5ex]
Relevance smoothness ($s$) & 0.001 & 0.1 \\[0.5ex]
\hline
\end{tabular}
\vspace{-0.3cm}
\end{table}

\subsection{Topology}
\label{sec:topology-eval}

In the proposed method, an adequate positioning of the topological nodes is vital for an adequate consolidation of the visual features. Thus, we evaluated the positioning of the nodes by checking if they were coherent with the spatial positions provided in the input data. For this, the data sequences that were captured in the same area were presented randomly to the method and a single map was generated per area, where each map was generated with 6 data sequences of 2 paths. The selected data contains a single area (paths 1 and 2) in COLD-Freiburg, and in COLD-Saarbrucken it contains area 1 (paths 1 and 2) and area 2 (paths 3 and 4). 

\begin{figure}[ht]
\centering
  \includegraphics[width=1.0\linewidth]{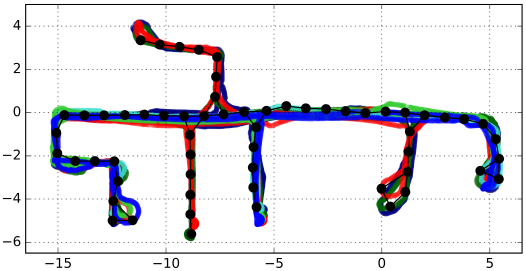}
  \caption{Nodes of a topological map generated with the selected data sequences from the COLD-Freiburg dataset, provided to the proposed method in random order, plot over the positions of all the input data. Each colored line (there are six colors) represents the positions provided in a data sequence. The nodes of the map and their connections are in black.}
  \label{fig:mapa-topologico-freiburg}
  \vspace{-0.4cm}
\end{figure}

The process was repeated 10 times per area and in each execution the positions of the nodes in the generated map were first visually evaluated by plotting them over the positions of all input data (Fig. \ref{fig:mapa-topologico-freiburg} shows a typical example). Then, the mean and maximum Euclidean distance between all input position data and their nearest nodes were computed. The results obtained in all executions were averaged, resulting in: $0.304 \pm 0.004$ / $0.873 \pm 0.010$ (mean / maximum distance) for Freiburg single area; $0.314 \pm 0.009$ / $0.863 \pm 0.020$ (mean / maximum distance) for Saarbrucken area 1; and, $0.313 \pm 0.022$ / $0.842 \pm 0.025$ (mean / maximum distance) for Saarbrucken area 2. 

The results allowed us to conclude that all the nodes were positioned coherently with the input data (with mean distance close to 0.3) and the spatial distance threshold $\lambda$ (with maximum distance always below $\lambda=0.9$). We also analyzed all connections between the nodes and verified that they all represented viable transitions in the environments.

\subsection{Object Classification}
\label{sec:objet-classification-eval}

Since the visual features of multiple images are consolidated in the same node $j$, the TOP-1 label assigned by the GoogleNet to an image may not be the TOP-1 label when classifying the consolidated visual features vector $\mathbf{c}_{j}$ using the same classification layer (as described in Section~\ref{sec:semantic-information}). However, we expect it to be ranked close to the top. To evaluate this, we compute how frequently the TOP-1 classification label (among all the 1000 classes in ImageNet\footnote{Most classes in ImageNet are objects, but they also include some classes of animals.}) of each image is present in the TOP-5 classification labels of $\mathbf{c}_{j}$ of each node $j$ that consolidates the image. Images can be consolidated by more than one node as long as they have been captured at an intersection position between nodes.

This evaluation is done per sub-dataset, with all selected data sequences from all paths and one map generated per data sequence. Tab.~\ref{table:tableResultObjectsClassification} presents the accuracy results obtained, which suggest that the consolidated visual features fairly preserve the classifications observed in the original images.

As the TOP-5 accuracy is not sensitive enough to detect small variations, to evaluate the effect of the visual persistence (VP) capability of the method, we classify the vector $\mathbf{c}_{j}$ of each node $j$ with the linear classification layer of the GoogLeNet, provide the outputs of $13$ classes of objects from ImageNet (selected based on what is found in the COLD images\footnote{Object classes selected for the evaluation: washbasin, soap dispenser, toilet seat, photocopier, monitor, desktop computer, desk, dining table, barber chair, microwave oven, stove, dishwasher and toaster.}) for a softmax layer and obtain the classification vector $\mathbf{o}_{j}$. Then, we obtain the average classification of all images (classified by GoogLeNet with the output of the same 13 classes provided for a softmax layer) captured around each node $j$ in the vector $\mathbf{m}_{j}$. Finally, we compare $\mathbf{m}_{j}$ to $\mathbf{o}_{j}$ for all nodes with an error measure based on the l1-norm: $err_{mean} = \tfrac{1}{b\times k}\sum_{j=1}^{k} ||o_j - m_j||_1$, where $b=13$ is the number of classes of objects we selected and $k$ assumes the number of nodes in all maps generated added. This evaluation is also done per sub-dataset, with one map generated per data sequence.

Tab.~\ref{table:tableResultObjectsClassification} presents the results obtained, which give evidence of the positive contribution of the visual features persistence in the consolidation process. The results also provide further evidence that the consolidated visual features are representative of their regions in order to approximate the classification results to those of the average classification of all images captured in those regions.

\begin{table}[ht]
\centering
\caption{Results of the objects classification evaluation.}
\label{table:tableResultObjectsClassification}
\begin{threeparttable}
\begin{tabular}{p{2.6cm}C{2.2cm}C{2.3cm}}
\hline
&\\[-2ex]
 & COLD-Freiburg & COLD-Saarbrucken \\[0.5ex]
\hline
&\\[-2ex]
$PM_{}$\tnote{*} $(Accuracy)$ & 0.8185 & 0.7616 \\ [0.5ex]
\hline
&\\[-1.8ex]
$PM_{}$ $(err_{mean})$ & 0.0123(0.0068) & 0.0163(0.0079) \\ [0.5ex]
$PM_{-VP}$ $(err_{mean})$ & 0.0146(0.0083) & 0.0196(0.0109) \\ [0.5ex]
\hline
\end{tabular}
\begin{tablenotes}\footnotesize
\item Standard deviations are shown in parentheses.
\item[*] $PM_{}$ is Proposed Method and $PM_{-VP}$ is Proposed Method without Visual Persistence.
\end{tablenotes}
\end{threeparttable}
\vspace{-0.4cm}
\end{table}

\begin{table*}[ht]
\centering
\caption{Results of the place classification experiment. Standard deviations are shown in parentheses.}
\label{table:tableResultPlaceClassificationPerPlaceCategory}
\begin{threeparttable}
\begin{tabular}{p{1.8cm}C{1.6cm}C{1.6cm}C{1.6cm}|C{1.6cm}C{1.6cm}C{1.6cm}}
\hline
&\\[-2ex]
\multirow{2}{*}{Accuracy} & \multicolumn{3}{c}{$PM_{}$\tnote{1}} & \multicolumn{3}{c}{\textit{IMAGES}} \\[0.5ex]
 & Fr1 and Fr2\tnote{2} & Sa1 and Sa2 & Sa3 and Sa4 & Fr1 and Fr2 & Sa1 and Sa2 & Sa3 and Sa4 \\[0.5ex]
\hline
&\\[-2ex]
Corridor & 0.987(0.018) & 0.984(0.016) & 1.000(0.000) & 0.975(0.011) & 0.958(0.021) & 0.964(0.007) \\ [0.5ex]
Printer area & 1.000(0.000) & 1.000(0.000) & 1.000(0.000) & 0.914(0.057) & 0.908(0.072) & 0.972(0.025) \\ [0.5ex]
Robotics lab & ---\tnote{3} & 0.944(0.078) & --- & --- & 0.665(0.013) & --- \\ [0.5ex]
Stairs area & 0.944(0.124) & --- & --- & 0.911(0.032) & --- & --- \\ [0.5ex]
Bathroom & 1.000(0.000) & 1.000(0.000) & 1.000(0.000) & 0.972(0.025) & 0.965(0.012) & 0.978(0.009) \\ [0.5ex]
Kitchen & 1.000(0.000) & --- & 0.970(0.043) & 0.730(0.147) & --- & 0.914(0.056) \\ [0.5ex]
Terminal room & --- & 0.968(0.022) & --- & --- & 0.917(0.071) & --- \\ [0.5ex]
Conference room & --- & 1.000(0.000) & --- & --- & 0.921(0.065) & --- \\ [0.5ex]
1-person office & 1.000(0.000) & 0.933(0.094) & 1.000(0.000) & 0.845(0.091) & 0.818(0.031) & 0.977(0.021) \\ [0.5ex]
2-person office & 0.907(0.099) & 0.917(0.186) & --- & 0.891(0.104) & 0.861(0.054) & --- \\ [0.5ex]
Large office & 1.000(0.000) & --- & --- & 0.699(0.072) & --- & --- \\ [0.5ex]
\hline
\\[-2ex]
Overall & 0.979(0.021) & 0.982(0.013) & 0.995(0.011) & 0.931(0.025) & 0.928(0.032) & 0.967(0.015) \\ [0.5ex]
\hline
\end{tabular}
\begin{tablenotes}\footnotesize
\item[1] $PM_{}$ is Proposed Method and \textit{IMAGES} denotes the results obtained with the direct classification of the visual features vectors extracted from the images.
\item[2] Fr(n) is Freiburg (n = path 1 or 2) and Sa(n) is Saarbrucken (n = path 1, 2, 3 or 4).
\item[3] --- means the place category is not present in the paths used.
\end{tablenotes}
\end{threeparttable}
\vspace{-0.5cm}
\end{table*}

\subsection{Place Classification}
\label{sec:place-classification-eval}

In this experiment, the consolidated visual features vector $\mathbf{c}_{j}$ of each node $j$ is feed into the MLP described in Section \ref{sec:semantic-information} to classify the nodes in one of the 11 place categories present in the dataset. This experiment is performed using an adaptation of the k-fold cross-validation method, in which each fold is a sequence of data selected from the datasets, $k$ is the number of sequences chosen, and the order of the sequences is randomly selected. Thus, all images in the $k$-1 sequences of data are used to train the MLP in batch mode, with all images per epoch. The number of epochs is experimentally determined and only one value is defined for each cross-validation run. The remaining data sequence is used in the generation of the map and the resulting consolidated visual features vector $\mathbf{c}_{j}$ of each node $j$ is classified by the MLP. Therefore, in this experiment we use per execution only the selected data sequences that were captured in the two paths of the same area.

\begin{figure*}[ht]
\centering
  \includegraphics[width=0.65\linewidth]{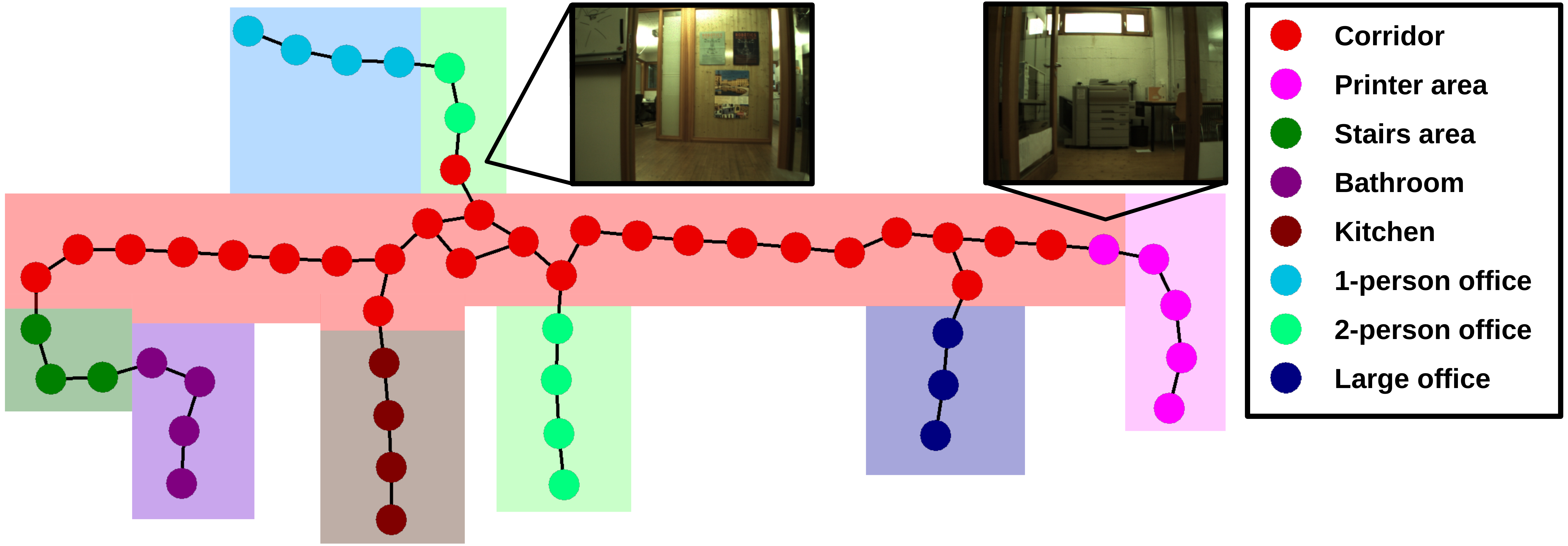}
  \caption{Example of place classification results obtained with the map generated from a path 2 data sequence of COLD-Freiburg. The color of each node is the class assigned by the MLP and the colored blocks represent the place categories of the nodes covered by the blocks in the ground truth. In the black boxes are examples of images captured in the regions of each misclassified node, these images demonstrate how visible is specific visual data of other place categories (eg., a printer at the entrance to the printer area) from the misclassified nodes.}
  \label{fig:mapa-topologico-freiburg-classes}
  \vspace{-0.3cm}
\end{figure*}

Tab.~\ref{table:tableResultPlaceClassificationPerPlaceCategory} presents the excellent results obtained (classification accuracy), which suggest that even a simple MLP can be used to accurately recognize the place categories of the nodes from the visual features consolidated in them. Fig.~\ref{fig:mapa-topologico-freiburg-classes} illustrates an example of the place classification results obtained with the map generated from a data sequence of COLD-Freiburg, in which we can see that only two nodes were misclassified. Both nodes are located close to the borders between different place categories in the ground truth, from where it is possible to see the other regions (Fig.~\ref{fig:mapa-topologico-freiburg-classes} exemplifies images), what may have led to the errors.

For comparison, we used the same adaptation of the cross-validation method with the difference that, at each run, the visual features vectors extracted from all images of the remaining data sequence are directly classified by the MLP. Again, the number of training epochs was experimentally determined and only one value was defined for each cross validation run. The results obtained (\textit{IMAGES} columns in Tab.~\ref{table:tableResultPlaceClassificationPerPlaceCategory}) were all smaller than the results obtained with the proposed method ($PM_{}$), which suggests that the proposed visual features consolidation method contributes positively to the classification of place categories. 

In addition, we performed a complementary comparison with the results of place classification obtained by \cite{rubio2016} and \cite{mancini2017}. Both works use the COLD dataset in a stratified 5-fold cross-validation process using only the path 2 images of each sub-dataset, separately. They consider each room as a place category and as there are two different 2-person office rooms in the path 2 of Freiburg, we use 9 categories for the Freiburg path 2 in this comparison only. The number of place categories used for the Saarbrucken path 2 remains 8. In \cite{rubio2016}, the authors extract HOG (Histogram of Oriented Gradients) features from 2D images and perform the classification using different classifiers. The best results from \cite{rubio2016} were obtained using a Bayesian Network (BN) as classifier and we used these results for comparison. \cite{mancini2017} introduces an end-to-end trainable deep approach to place classification that integrates a NBNN (Naive Bayes Nearest Neighbor) model into a CNN.

In order to approximate the experimentation scenarios for comparison, we use the same cross validation adaptation used earlier with the proposed method and consider only the selected data from path 2 of each sub-dataset. Tab.~\ref{table:tableResultPlaceClassificationComparison} presents the results obtained (denoted here as room classification), which are higher than the best obtained by \cite{rubio2016} and similar to the results of \cite{mancini2017}. This suggests that the features consolidated with the proposed method can also be used to recognize the specific room of the topological nodes with similar accuracy to the current literature.

\begin{table}[ht]
\centering
\caption{Room classification comparison with \cite{rubio2016} and \cite{mancini2017}.}
\label{table:tableResultPlaceClassificationComparison}
\begin{threeparttable}
\begin{tabular}{p{2.2cm}C{2.3cm}C{2.3cm}}
\hline
&\\[-2ex]
Accuracy & Fr2 & Sa2 \\[0.5ex]
\hline
&\\[-2ex]
Rubio et. al. \cite{rubio2016} & 0.823\tnote{1} & 0.844 \\ [0.5ex]
Mancini et. al. \cite{mancini2017} & 0.952 & 0.973 \\ [0.5ex]
$PM_{}$ & 0.957(0.018) & 0.973(0.011) \\ [0.5ex]
\hline
\end{tabular}
\begin{tablenotes}\footnotesize
\item Standard deviations (STD) are shown in parentheses.
\item[1] \cite{rubio2016} and \cite{mancini2017} did not provide variance or STD values.
\end{tablenotes}
\end{threeparttable}
\vspace{-0.5cm}
\end{table}

\begin{figure*}[ht]
\centering
\subfigure[]{
  \includegraphics[width=0.649\columnwidth]{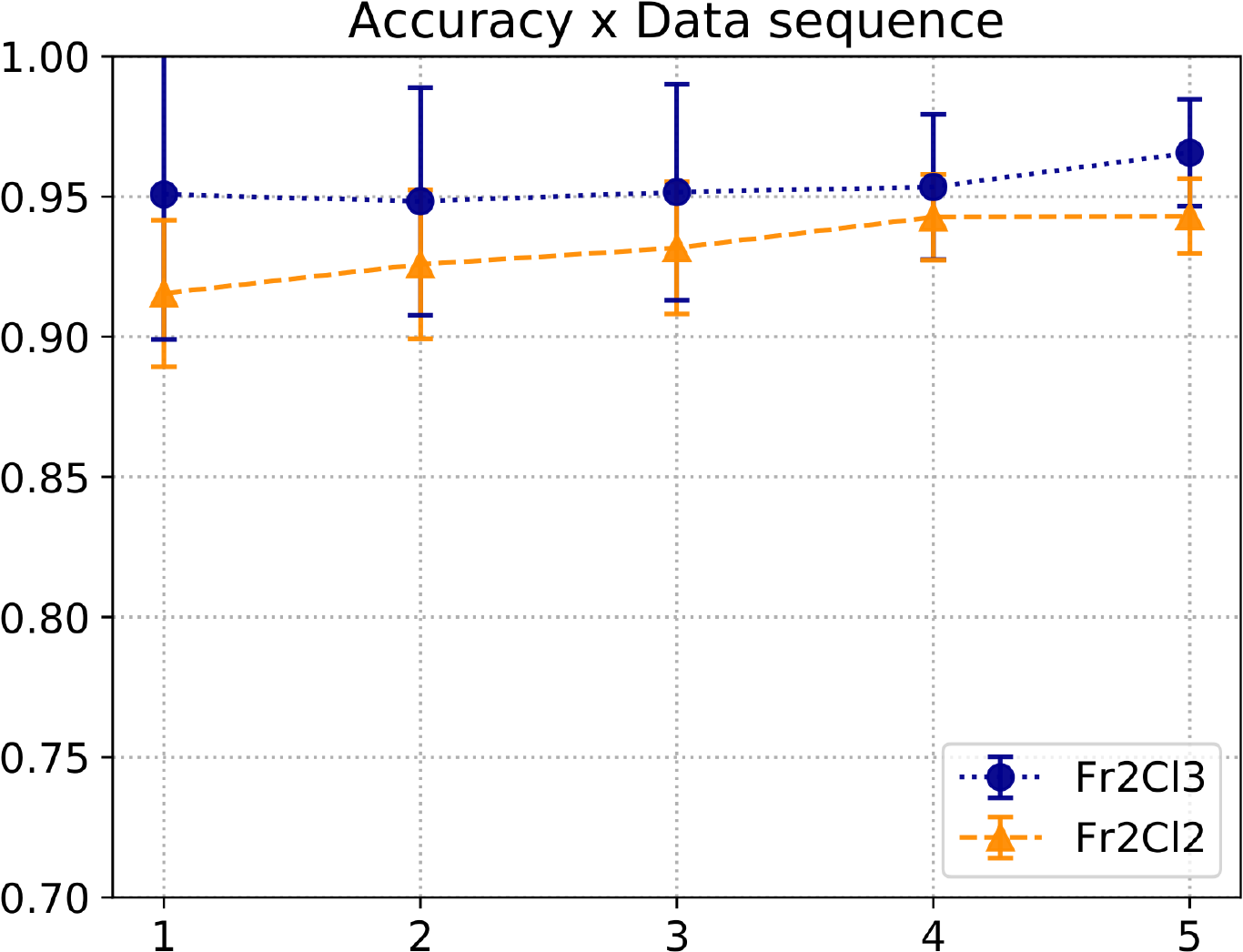}
  \label{subfig:place-classification-over-time-a}}
\subfigure[]{
  \includegraphics[width=0.649\columnwidth]{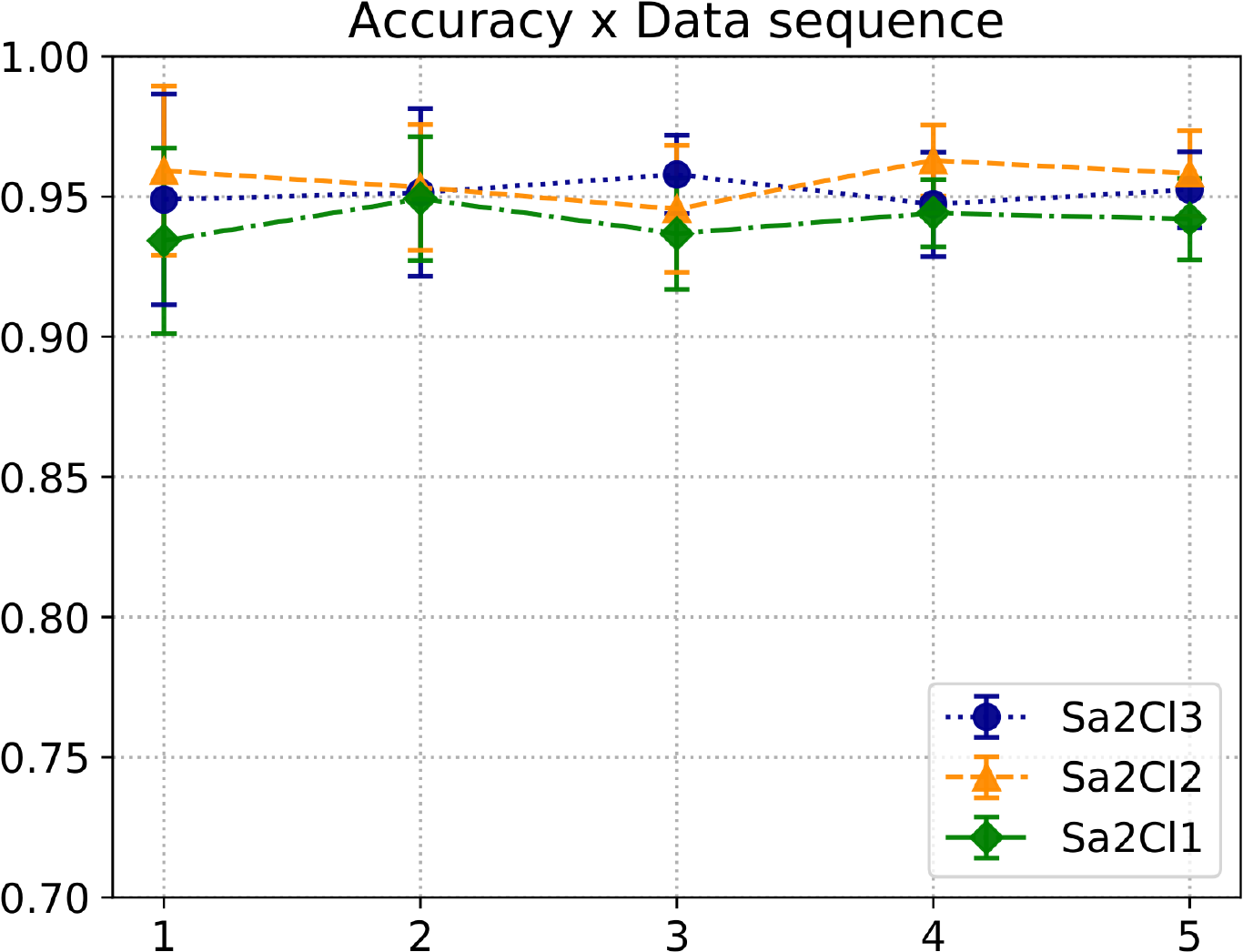}
  \label{subfig:place-classification-over-time-b}}
\subfigure[]{
  \includegraphics[width=0.649\columnwidth]{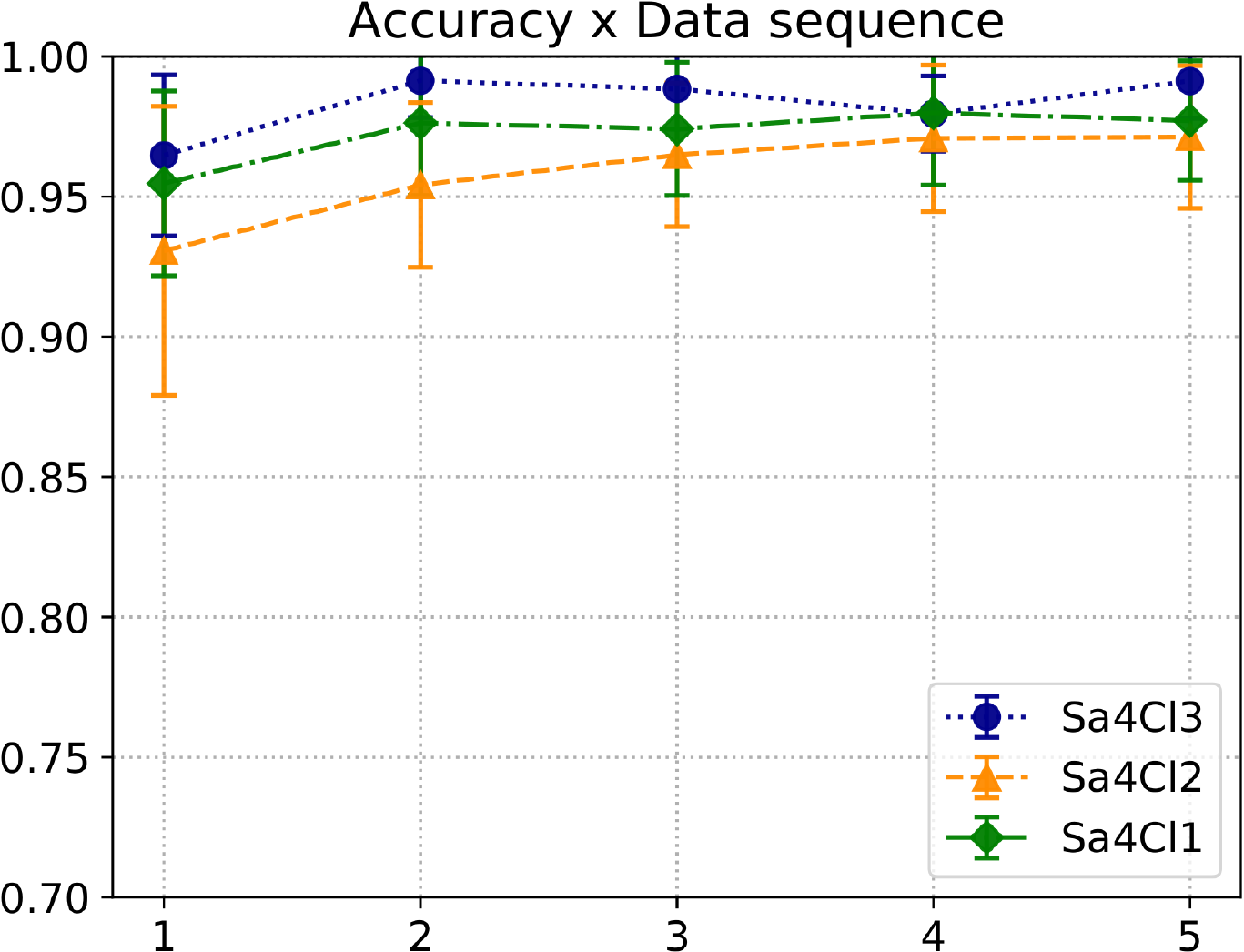}
  \label{subfig:place-classification-over-time-c}}
\caption{Results of the place classification over 5 instants of time: (a) shows the average results obtained with the data from Freiburg area; (b) and (c) show the average results obtained with the data from Saarbrucken area 1 and 2, respectively. Each legend item is the data sequence used to train the MLP. Fr2Cl1 was not used in (a) because it does not contain all the place categories.} 
\label{fig:place-classification-over-time}
\vspace{-0.5cm}
\end{figure*}

\subsection{Place Classification Over Time}
\label{sec:place-classification-over-time-eval}

In this experiment, from each area present in the dataset, we first selected data sequences of the longest path that contain all the place categories found in that respective area. Then, in an evaluation procedure repeated 10 times, we use one of these data sequences to train the MLP and the remaining 5 data sequences from the same area are used to generate the map. In the generation process, when all data in a data sequence is finished, the current state of the consolidated features vector $\mathbf{c}_{j}$ of each node $j$ is classified by the MLP. Thus, the evaluation takes place in 5 instants of time in the mapping process. This evaluation procedure is performed with each data sequence selected from the longest paths and the order of the remaining data sequences used in the generation of the maps is randomized at each run. 

The mean and standard deviation of the accuracy results obtained at each instant of time are illustrated in the graphs in Fig. \ref{fig:place-classification-over-time}. They show that for all three areas the results are stable or increasing as more data is consolidated in the mapping process. This suggests that the consolidated representations do not degrade with more data or might even improve. Fig. \ref{fig:place-classification-over-time} also shows that the standard deviations tend to get smaller as more data is consolidated into the maps, suggesting that the consolidated representations are less biased in relation to the initial features received and more representative of the regions covered by each topological node.

\subsection{Image Localization Experiment}
\label{sec:loc-evaluation}

\begin{table*}[ht]
\centering
\caption{Results of the image localization experiment with random image replication. Standard deviations are shown in parentheses. 
}
\label{table:tableResultImageLocalization2}
\begin{threeparttable}
\begin{tabular}{C{0.2cm}C{4.35cm}C{1.5cm}C{1.5cm}C{1.5cm}C{1.5cm}}
\hline
&&\\[-2ex]
\multirow{2}{*}{IR\tnote{1}} & \multirow{2}{*}{Data Sequences} & \multicolumn{2}{c}{$PM_{}$\tnote{2}} & \multicolumn{2}{c}{$PM_{-VH}$} \\[0.5ex]
& & TOP-1 Acc & TOP-5 Acc & TOP-1 Acc & TOP-5 Acc \\[0.5ex]
\hline
&&\\[-1.5ex]
\multirow{3}{*}{0} & Fr1(Cl1,Nt1,Nt3), Fr2(Cl1,Cl2,Cl3)\tnote{3} & 0.744(0.067) & 0.907(0.033) & 0.741(0.036) & 0.912(0.014) \\ [0.5ex]
& Sa1(Cl1,Cl2,Cl3), Sa2(Cl1,Cl2,Cl3) & 0.697(0.059) & 0.911(0.028) & 0.700(0.072) & 0.907(0.034) \\ [0.5ex]
& Sa3(Cl1,Cl2,Cl3), Sa4(Cl1,Cl2,Cl3) & 0.834(0.021) & 0.976(0.016) & 0.837(0.015) & 0.973(0.017) \\ [0.5ex]
\hline
&&\\[-1.5ex]
\multirow{3}{*}{20} & Fr1(Cl1,Nt1,Nt3), Fr2(Cl1,Cl2,Cl3) & 0.746(0.066) & 0.904(0.019) & 0.686(0.076) & 0.883(0.054) \\ [0.5ex]
& Sa1(Cl1,Cl2,Cl3), Sa2(Cl1,Cl2,Cl3) & 0.693(0.065) & 0.918(0.037) & 0.672(0.076) & 0.893(0.043) \\ [0.5ex]
& Sa3(Cl1,Cl2,Cl3), Sa4(Cl1,Cl2,Cl3) & 0.846(0.024) & 0.976(0.018) & 0.790(0.040) & 0.968(0.021) \\ [0.5ex]
\hline
\\[-1.5ex]
\multirow{3}{*}{40} & Fr1(Cl1,Nt1,Nt3), Fr2(Cl1,Cl2,Cl3) & 0.759(0.060) & 0.902(0.020) & 0.638(0.068) & 0.880(0.030) \\ [0.5ex]
& Sa1(Cl1,Cl2,Cl3), Sa2(Cl1,Cl2,Cl3) & 0.693(0.052) & 0.913(0.030) & 0.650(0.079) & 0.882(0.042) \\ [0.5ex]
& Sa3(Cl1,Cl2,Cl3), Sa4(Cl1,Cl2,Cl3) & 0.833(0.043) & 0.976(0.016) & 0.753(0.061) & 0.951(0.017) \\ [0.5ex]
\hline
\end{tabular}
\begin{tablenotes}\footnotesize
\item[1] IR is nº of image replications and Acc stands for Accuracy.
\item[2] $PM_{}$ is Proposed Method and $PM_{-VH}$ is Proposed Method without Visual Habituation.
\item[3] Fr(n) is Freiburg (n = path 1 or 2), Sa(n) is Saarbrucken (n = path 1, 2, 3 or 4), Cl(n) is cloudy sequence (n = sequence 1, 2 or 3) and Nt(n) is night sequence (n = sequence 1 or 3).
\end{tablenotes}
\end{threeparttable}
\vspace{-0.5cm}
\end{table*}

In the last experiment, we use the visual data consolidated in each node to localize images on the maps using the approach described in Section \ref{sec:image-localization} and study the effect of the visual habituation (VH) capability of the method. This experiment is performed using the same adaptation of the k-fold cross-validation method used earlier with the proposed method in Section \ref{sec:place-classification-eval}, but here $k$-1 sequences of data are used in the generation of the map and all images in the remaining sequence of data are localized on the map. Thus, in this experiment we use per execution only the selected data sequences that were captured in the same area and a single map is generated per area.

We evaluated this in three scenarios in which 10\% of the images in each sequence were replicated 0 times in the first scenario, 20 in the second, and 40 times in the third. The images to be replicated were randomly selected at the beginning of the map construction in each execution from the data sequences used to build the map. With this, we simulate moments when the robot stops moving and captures very similar images for a while. Tab.~\ref{table:tableResultImageLocalization2} shows the TOP-1 and TOP-5 accuracy values measured with the proposed method with ($PM_{}$) and without visual habituation capability ($PM_{-VH}$). TOP-1 evaluates whether for each image the node on the map with the highest activation is the node (or one of the nodes) in the ground truth. TOP-5 evaluates whether, for each image, one of the five nodes on the map with the highest activation is the node (or one of the nodes) in the ground truth. The ground truth for an image can be more than one node, if it was captured at an intersection position between nodes.

The results with the complete method ($PM_{}$) remained stable as the number of image replications increased, which shows the positive influence of the visual habituation capability in the consolidation of visual features. While the results in $PM_{-VH}$ got worse as the number of image replications increased, which suggests that without the visual habituation capability it would increasingly degrade its consolidated features the longer the robot stays put, due to the sequential presentation of similar features. In addition, the results in $PM_{}$ suggest that the consolidated representations in each node can be used to locate images with good accuracy. 

\section{CONCLUSIONS}
\label{sec:conclusion}

This paper presented a topological semantic mapping method that consolidates visual features of regions. The consolidated features allowed flexible classification of semantic properties (objects, room, and place category) of the regions covered by the nodes and use in other visual tasks (image localization) with very promising results in indoor environments. To evaluate further the applicability of the model, in future work we will assess it with clustering methods and new classifiers trained on images from other environments, in order to recognize new semantic properties of previously generated maps. 

\addtolength{\textheight}{-0cm}








\begin{thebibliography}{10}
\providecommand{\url}[1]{#1}
\csname url@rmstyle\endcsname
\providecommand{\newblock}{\relax}
\providecommand{\bibinfo}[2]{#2}
\providecommand\BIBentrySTDinterwordspacing{\spaceskip=0pt\relax}
\providecommand\BIBentryALTinterwordstretchfactor{4}
\providecommand\BIBentryALTinterwordspacing{\spaceskip=\fontdimen2\font plus
\BIBentryALTinterwordstretchfactor\fontdimen3\font minus
  \fontdimen4\font\relax}
\providecommand\BIBforeignlanguage[2]{{%
\expandafter\ifx\csname l@#1\endcsname\relax
\typeout{** WARNING: IEEEtran.bst: No hyphenation pattern has been}%
\typeout{** loaded for the language `#1'. Using the pattern for}%
\typeout{** the default language instead.}%
\else
\language=\csname l@#1\endcsname
\fi
#2}}

\bibitem{kostavelis2015}
I.~Kostavelis and A.~Gasteratos, ``Semantic mapping for mobile robotics tasks:
  A survey,'' \emph{Robotics and Autonomous Systems}, vol.~66, pp. 86 -- 103,
  2015.

\bibitem{pronobis2012}
A.~Pronobis and P.~Jensfelt, ``Large-scale semantic mapping and reasoning with
  heterogeneous modalities,'' in \emph{International Conference on Robotics and
  Automation (ICRA)}.\hskip 1em plus 0.5em minus 0.4em\relax IEEE, 2012, pp.
  3515--3522.

\bibitem{ma2017}
L.~{Ma}, J.~{Stückler}, C.~{Kerl}, and D.~{Cremers}, ``Multi-view deep
  learning for consistent semantic mapping with {RGB-D} cameras,'' in
  \emph{IEEE/RSJ International Conference on Intelligent Robots and Systems
  (IROS)}, 2017, pp. 598--605.

\bibitem{xiang2017}
Y.~Xiang and D.~Fox, ``{DA-RNN}: Semantic mapping with data associated
  recurrent neural networks,'' in \emph{Robotics: Science and Systems},
  Cambridge, Massachusetts, July 2017.

\bibitem{sunderhauf2016}
N.~Sunderhauf, F.~Dayoub, S.~Mcmahon, B.~Talbot, R.~Schultz, P.~Corke,
  G.~Wyeth, B.~Upcroft, and M.~Milford, ``Place categorization and semantic
  mapping on a mobile robot,'' in \emph{International Conference on Robotics
  and Automation (ICRA)}, May 2016, pp. 5729--5736.

\bibitem{roddick2020}
T.~Roddick and R.~Cipolla, ``Predicting semantic map representations from
  images using pyramid occupancy networks,'' in \emph{IEEE/CVF Conference on
  Computer Vision and Pattern Recognition (CVPR)}, 2020.

\bibitem{mccormac2017}
J.~{McCormac}, A.~{Handa}, A.~{Davison}, and S.~{Leutenegger},
  ``Semanticfusion: Dense {3D} semantic mapping with convolutional neural
  networks,'' in \emph{IEEE International Conference on Robotics and Automation
  (ICRA)}, 2017, pp. 4628--4635.

\bibitem{maturana2017}
D.~{Maturana}, S.~{Arora}, and S.~{Scherer}, ``Looking forward: A semantic
  mapping system for scouting with micro-aerial vehicles,'' in \emph{IEEE/RSJ
  International Conference on Intelligent Robots and Systems (IROS)}, 2017, pp.
  6691--6698.

\bibitem{maturana2018}
D.~Maturana, P.-W. Chou, M.~Uenoyama, and S.~Scherer, ``Real-time semantic
  mapping for autonomous off-road navigation,'' in \emph{Field and Service
  Robotics}, M.~Hutter and R.~Siegwart, Eds.\hskip 1em plus 0.5em minus
  0.4em\relax Springer International Publishing, 2018, pp. 335--350.

\bibitem{bernuy2018}
F.~Bernuy and J.~Ruiz-del Solar, ``Topological semantic mapping and
  localization in urban road scenarios,'' \emph{Journal of Intelligent \&
  Robotic Systems}, vol.~92, no.~1, pp. 19--32, 2018.

\bibitem{luo2018}
R.~C. {Luo} and M.~{Chiou}, ``Hierarchical semantic mapping using convolutional
  neural networks for intelligent service robotics,'' \emph{IEEE Access},
  vol.~6, pp. 61\,287--61\,294, 2018.

\bibitem{sunderhauf2017}
N.~{Sünderhauf}, T.~T. {Pham}, Y.~{Latif}, M.~{Milford}, and I.~{Reid},
  ``Meaningful maps with object-oriented semantic mapping,'' in \emph{IEEE/RSJ
  International Conference on Intelligent Robots and Systems (IROS)}, 2017, pp.
  5079--5085.

\bibitem{rangel2019}
J.~C. {Rangel}, M.~{Cazorla}, I.~{Garc{\'\i}a-Varea}, C.~{Romero-Gonz{\'a}lez},
  and J.~{Mart{\'\i}nez-G{\'o}mez}, ``Automatic semantic maps generation from
  lexical annotations,'' \emph{Autonomous Robots}, vol.~43, no.~3, pp.
  697--712, 2019.

\bibitem{sousa2018}
Y.~C.~N. Sousa and H.~F. Bassani, ``Incremental semantic mapping with
  unsupervised on-line learning,'' in \emph{International Joint Conference on
  Neural Networks (IJCNN)}, July 2018.

\bibitem{nakajima2019}
Y.~{Nakajima} and H.~{Saito}, ``Efficient object-oriented semantic mapping with
  object detector,'' \emph{IEEE Access}, vol.~7, pp. 3206--3213, 2019.

\bibitem{rosinol2021}
A.~Rosinol, A.~Violette, M.~Abate, N.~Hughes, Y.~Chang, J.~Shi, A.~Gupta, and
  L.~Carlone, ``Kimera: from {SLAM} to spatial perception with {3D} dynamic
  scene graphs,'' \emph{The International Journal of Robotics Research}, 2021.

\bibitem{szegedy2015}
C.~Szegedy, W.~Liu, Y.~Jia, P.~Sermanet, S.~Reed, D.~Anguelov, D.~Erhan,
  V.~Vanhoucke, and A.~Rabinovich, ``Going deeper with convolutions,'' in
  \emph{IEEE Conference on Computer Vision and Pattern Recognition (CVPR)},
  June 2015.

\bibitem{bassani2015}
H.~F. Bassani and A.~F.~R. Araujo, ``Dimension selective self-organizing maps
  with time-varying structure for subspace and projected clustering,''
  \emph{IEEE Transactions on Neural Networks and Learning Systems}, vol.~26,
  no.~3, pp. 458--471, 2015.

\bibitem{pronobis2009}
A.~Pronobis and B.~Caputo, ``{COLD}: The {CoSy} localization database,''
  \emph{The International Journal of Robotics Research}, vol.~28, no.~5, 2009.

\bibitem{araujo2013}
A.~F. Araujo and R.~L. Rego, ``Self-organizing maps with a time-varying
  structure,'' \emph{ACM Comput. Surv.}, vol.~46, no.~1, July 2013.

\bibitem{imagenet2009}
\BIBentryALTinterwordspacing
J.~Deng, W.~Dong, R.~Socher, L.-J. Li, K.~Li, and L.~Fei-Fei, ``{ImageNet: A
  Large-Scale Hierarchical Image Database},'' 2009. [Online]. Available:
  \url{http://image-net.org/}
\BIBentrySTDinterwordspacing

\bibitem{kingma2014}
D.~P. Kingma and J.~Ba, ``Adam: A method for stochastic optimization,''
  \emph{arXiv preprint arXiv:1412.6980}, 2014.

\bibitem{bassani2012}
H.~F. Bassani and A.~F.~R. Araujo, ``Dimension selective self-organizing maps
  for clustering high dimensional,'' in \emph{International Joint Conference on
  Neural Networks (IJCNN)}.\hskip 1em plus 0.5em minus 0.4em\relax IEEE, June
  2012.

\bibitem{Helton2005}
J.~C. Helton, F.~J. Davis, and J.~D. Johnson, ``A comparison of uncertainty and
  sensitivity analysis results obtained with random and latin hypercube
  sampling.'' \emph{Reliability Engineering and System Safety}, vol.~89, pp.
  305--330, 2005.

\bibitem{rubio2016}
F.~Rubio, J.~Martínez-Gómez, M.~{Julia Flores}, and J.~M. Puerta,
  ``Comparison between bayesian network classifiers and {SVMs} for semantic
  localization,'' \emph{Expert Systems with Applications}, 2016.

\bibitem{mancini2017}
M.~Mancini, S.~R. Bulò, E.~Ricci, and B.~Caputo, ``Learning deep {NBNN}
  representations for robust place categorization,'' \emph{IEEE Robotics and
  Automation Letters}, vol.~2, no.~3, pp. 1794--1801, 2017.

\end{thebibliography}

\end{document}